\documentclass[sigconf]{acmart}

\makeatletter
\@ifpackageloaded{graphicx}{}{\usepackage{graphicx}}
\makeatother
\usepackage[ruled,vlined,linesnumbered]{algorithm2e}
\DontPrintSemicolon
\usepackage{makecell, multirow, dcolumn}
\usepackage{colortbl}
\newcommand{\shade}{\cellcolor{gray!15}}
\makeatletter
\@ifpackageloaded{framed}{}{\usepackage{framed}}
\makeatother
\definecolor{shadecolor}{RGB}{250, 245, 235}
\newenvironment{insightbox}
  {
   \MakeFramed{\hsize=0.93\linewidth \FrameRestore}
   \raggedright}
  {\endMakeFramed}
\usepackage{pifont}
\usepackage{xspace}
\newcommand{\momo}{MoMoVL\xspace}
\newcommand{\moW}{MoMoVL-Weight\xspace}
\newcommand{\moI}{MoMoVL-Index\xspace}

\copyrightyear{2027}
\acmYear{2027}
\setcopyright{cc}
\setcctype{by}
\acmConference[WSDM '27]{Proceedings of the Twentieth ACM International Conference on Web Search and Data Mining}{February 15--19, 2027}{Hong Kong, Hong Kong}
\acmBooktitle{Proceedings of the Twentieth ACM International Conference on Web Search and Data Mining (WSDM '27), February 15--19, 2027, Hong Kong, Hong Kong}
\acmDOI{10.1145/3862322.3863775}
\acmISBN{979-8-4007-3117-4/2027/02}

\begin{document}

\title{MoMoVL: Modality-Aware Adaptive Compression for On-Device Sparsely-Activated Vision-Language Models}

\author{Dongkeun Lee}
\affiliation{
  \institution{Korea University}
  \city{Seoul}
  \country{Republic of Korea}
}
\email{dklee98@korea.ac.kr}

\author{Wonjun Lee}
\affiliation{
  \institution{Gachon University}
  \city{Seongnam}
  \country{Republic of Korea}
}
\email{wonjun.lee@gachon.ac.kr}

\begin{abstract}
The rise of the agentic web has been accompanied by growing demand for on-device Vision-Language Models (VLMs) that perform diverse cross-modal web tasks with enhanced privacy and reduced latency. 
Mixture-of-Experts (MoE) models are widely adopted in modern VLMs for compute-efficient scalability, yet their multi-expert architecture poses deployment challenges due to substantial memory demands.
Mixed-precision quantization is an effective compression technique that reduces bit-widths while adapting to the heterogeneous hardware budgets of edge devices, particularly well-suited for MoE due to its inherent structural sparsity, which causes imbalanced importance among experts.
While prior work has applied quantization for efficient MoE inference at the edge, we point out that in MoE-based VLMs, the modality-specific characteristics of visual and text tokens further impact expert quantization.
To address this issue, we present MoMoVL, a modality-aware adaptive compression framework for MoE-based VLMs that brings vision-language inference within reach of on-device resource budgets.
Motivated by our empirical findings on modality-induced disparities across experts, MoMoVL jointly exploits the per-expert activation profile and per-modality quantization sensitivity to guide bit-width assignment.
Experimental results show that MoMoVL improves the accuracy-compression trade-off relative to state-of-the-art mixed-precision baselines and reduces memory footprint by up to 15.44\% with a modest drop in accuracy over its uniform-precision baseline.
Our code and data are publicly available at \url{https://github.com/sco-edge/MoMoVL}.
\end{abstract}

\ccsdesc[500]{Information systems~Multimedia information systems}

\keywords{Mixture-of-Experts, Vision-Language Models, Edge Computing}

\maketitle

\section{Introduction}
\label{sec:intro}

The rise of the agentic web, where web agents autonomously search, interact with, and perform tasks across diverse web applications, has been accompanied by increasing interest in on-device deployment of the underlying foundation models.
At the heart of web agents \cite{SEEACT, ColorBrowserAgent, litewebagent} are Vision-Language Models (VLMs), equipped with multimodal capabilities to understand webpage content and perform visual reasoning \cite{CLIP, Flamingo, llava}. 
While VLMs are being widely adopted due to their strong performance in handling various cross-modal tasks \cite{AGREE_WSDM23,MOON_WSDM26}, deploying them solely in the cloud raises concerns around privacy and data security \cite{Sim2023}, as well as latency and service cost \cite{ShiftVL}, driving growing demand for their on-device execution \cite{MiniCPM-V,afm3,gemini_nano}.
Recent studies on optimizing VLMs for edge environments have focused on model pruning \cite{EfficientVLM}, knowledge distillation \cite{distilling_NN}, quantization \cite{nagel2021}, and hardware acceleration \cite{hexagon_npu}.
Despite these efforts, these VLMs are yet to be practically deployed on edge devices not only due to resource-limited settings, but also the high heterogeneity in hardware budgets available per device.

Mixture-of-Experts (MoE) models are computationally well-suited for edge deployment.
Scaling up model capacity generally involves increasing model parameters, which leads to higher computation overhead and latency at inference.
Unlike traditional dense models, in which all parameters are activated for every input, MoE models activate only a subset of experts for each input token based on gating decisions \cite{Shazeer2017,gshard}.
This conditional computation enables them to achieve performance competitive with dense models of larger active parameter count, while requiring substantially fewer FLOPs per input \cite{Switch_transformers}.
The sublinear scaling of computation cost exhibited by MoE models has led to their wide adoption across various domains including information retrieval \cite{MoARE_WSDM26}, automatic speech recognition \cite{CMoE}, and vision-language tasks \cite{MolmoE, DeepSeek-VL2, afm3, kimivl}. 

The computational advantage of sparse activation, however, does not extend to memory. Because the multi-expert architecture requires all experts to remain resident in memory regardless of their activation status, the memory footprint scales with the total parameter count rather than the active one.
MolmoE-1B \cite{MolmoE}, for instance, activates roughly 1.5B parameters per token yet requires all 7.2B to be held in memory.
Such disparity grows with model scale, as the number of experts increases while the number activated per token remains relatively fixed.
Mitigating the resulting memory overhead requires compression techniques that reduce bit-widths while adapting to the differing resource budgets of target devices.

MoE's multi-expert, sparsely-activated design provides two unique opportunities for efficient quantization.
First, while expert weights account for the majority of model parameters, e.g., 89.21\% for MolmoE-1B, quantizing expert Feed-Forward Network (FFN) layers is far less damaging to model quality than quantizing other non-expert components, such as self-attention or dense layers \cite{MoQE}. The robustness of experts to quantization allows them to be aggressively compressed to 2 bits \cite{Fate, MoE-APEX}, or even to just 1 bit \cite{QMoE, CMoE}.
Second, the inherent structural sparsity causes imbalanced importance among experts, making mixed-precision quantization particularly effective. A small subset of ``hot'' experts is activated more frequently, contributing more to the model output. For example, the top 4 most frequently activated experts in MolmoE-1B account for 25.30\% of its total activation count. 
Previous studies have utilized these opportunities to address the deployment challenges in resource-constrained edge devices, focusing on expert quantization \cite{MoE-Offloading, MoE-CSP} and allocating different bit-widths to each expert based on its importance \cite{MC-MoE,Fate,MoE-APEX,EdgeMoE}.
Yet these approaches target a single-modality architecture, i.e., language-only MoE models.
In MoE-based VLMs, the modality-specific characteristics of visual and text tokens further impact expert quantization.

We find that in MoE-based VLMs, experts show imbalanced activation across modalities, leading to varying sensitivity to quantization.
It has been well studied that different experts tend to become highly specialized along the input structure \cite{Shazeer2017}, and our empirical analysis shows that in MoE-based VLMs, this further extends to the modality axis.
As shown in Fig. \ref{fig:fig2}, there exists a clear distinction between groups of \textbf{visual-centric} experts and \textbf{text-centric} experts (i.e., experts frequently activated when processing visual tokens and text tokens, respectively).
When quantized, this expert-level asymmetry is compounded by a token-level one.
Text tokens, despite being fewer in number (cf. Table \ref{tab:tab2}), carry substantially higher information density than spatially redundant visual tokens \cite{MBQ, MABA}, making text-centric experts more susceptible to quantization-induced degradation.
Details on these modality-driven expert asymmetries are discussed in Section \ref{sec:insight}.

Therefore, applying a unified quantization across these experts, ignoring both their relative importance and their dominant modality, leads to suboptimal results.
While existing studies propose various quantization strategies tailored for VLMs, only a few recent efforts \cite{MBQ,MABA,MASQuant,VEQ} begin to consider the varying sensitivity to quantization across modalities.
Of these, most target dense VLMs \cite{MBQ, MABA, MASQuant}, while VEQ \cite{VEQ} is the only such attempt that takes into account the structural properties of MoE by incorporating expert importance into the reconstruction loss.
However, it operates only under a fixed, uniform precision (3-bit or 4-bit), making it hard to fully exploit the available on-device budget, potentially sacrificing more of the underlying VLM's capabilities than necessary.
Conversely, MoPEQ \cite{MoPEQ} allocates bit-widths at the expert level based on their relative importance, but remains unaware of the modality-driven expert differentiation established above.
To the best of our knowledge, no prior work jointly assigns bit-widths across experts in a modality-aware manner while adapting to the heterogeneous, device-specific resource budgets of edge deployment.

To address this issue, we present \momo, a \underline{\textbf{Mo}}dality-aware adaptive compression framework for \underline{\textbf{Mo}}E-based \underline{\textbf{VL}}Ms that brings vision-language inference within reach of on-device resource budgets.
The key idea of \momo is to build a per-modality activation profile for each expert, measuring its activation statistics separately for visual and text tokens using a small multimodal calibration set.
In addition, \momo identifies how the impact of quantization noise diverges by modality and uses gradients of the Supervised Fine-Tuning (SFT) loss \cite{llava} as a proxy for the per-modality quantization sensitivity.
Finally, \momo reweights the activation profile by this sensitivity scale, yielding a combined importance metric per expert.
This balanced measure of expert importance enables \momo to perform modality-aware bit-width assignment subject to device-specific budget constraints, rather than statically quantizing the model to a fixed, uniform precision.
The major contributions of this work are summarized as follows:
\begin{itemize}
  \item We propose a novel modality-aware, adaptive compression framework tailored for on-device MoE-based VLMs that assigns bit-widths under device-specific hardware budgets. Unlike prior mixed-precision approaches, our method jointly exploits the structural properties and modality-dependent characteristics of multimodal MoE models to guide expert-wise bit-width assignment.  
  
  \item We present empirical results that demonstrate modality-induced disparities across experts in terms of activation statistics and quantization sensitivity, where overlooking either significantly degrades quantization performance. Our findings reveal that this performance is largely governed by an expert's dominant modality. Selectively quantizing text-centric experts at higher bit-width improves accuracy by up to 4.98\% over prioritizing visual-centric experts.
  
  \item We conduct comprehensive experiments on two popular MoE-based VLMs: MolmoE-1B and DeepSeek-VL2-Tiny, across eight widely used multimodal benchmarks covering diverse vision-language reasoning tasks. Our approach improves the accuracy-compression trade-off over state-of-the-art mixed-precision baselines and reduces memory footprint by up to 15.44\% with a modest drop in accuracy relative to its uniform-precision baseline.
\end{itemize}

\begin{figure}[t]
    \centering
    \includegraphics[width=.91\linewidth]{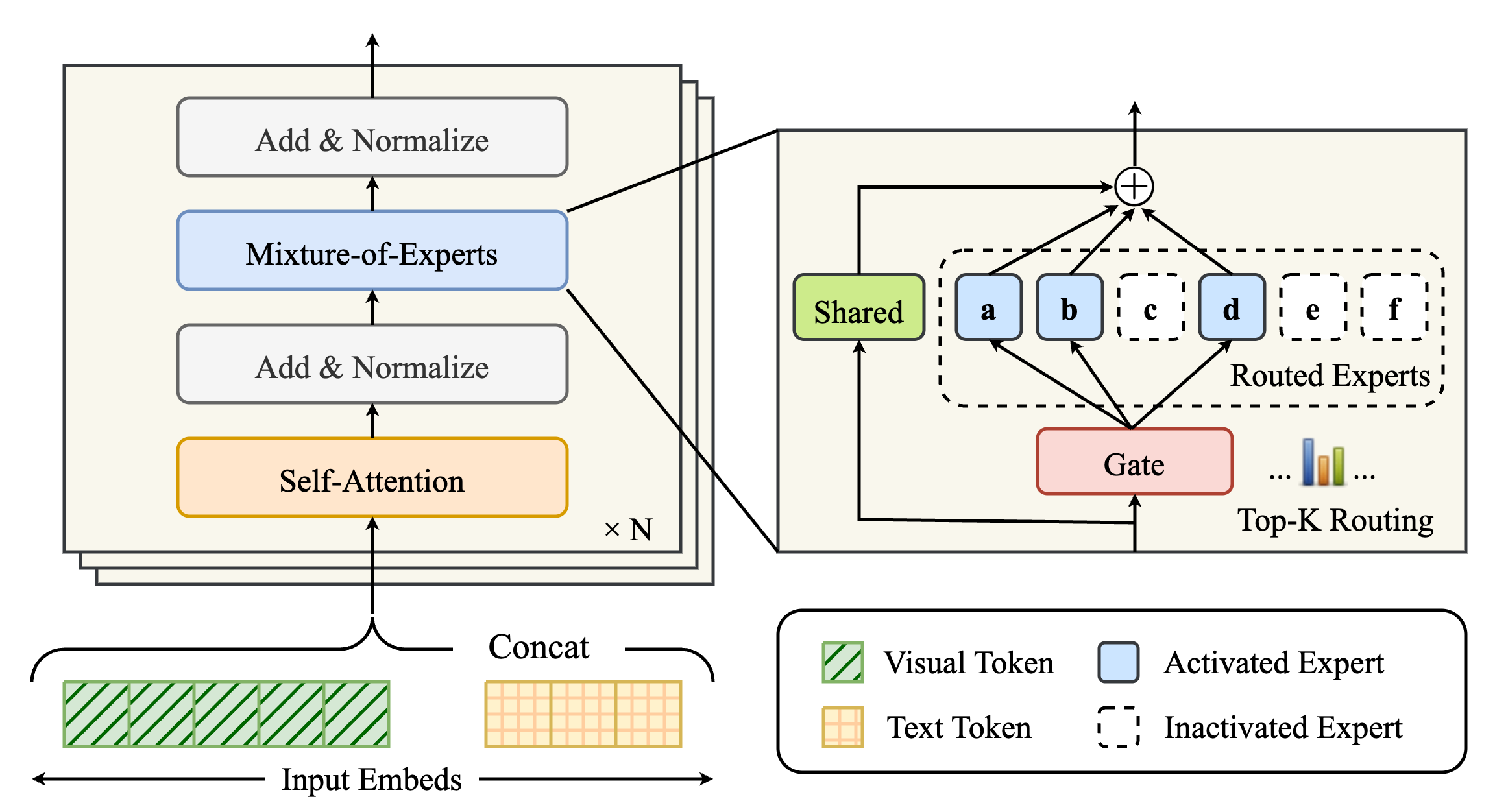}
    \Description[Overview of MoE-based VLM]{Input embeddings comprising visual and text tokens are processed with their routed experts based on gating weights, along with the shared experts that are always activated.}
    \caption{Architectural overview of an MoE-based VLM. \mbox{Visual} tokens and text tokens are processed with shared and routed experts using top-k routing.}
    \label{fig:fig1}
\end{figure}

\section{Background and Related Work}
\label{sec:bg}
In this section, we briefly review the mechanisms of MoE-based VLMs, followed by prior work on optimizing VLMs and MoE models for edge deployment.

\begin{figure*}[t]
    \centering
    \includegraphics[width=\textwidth]{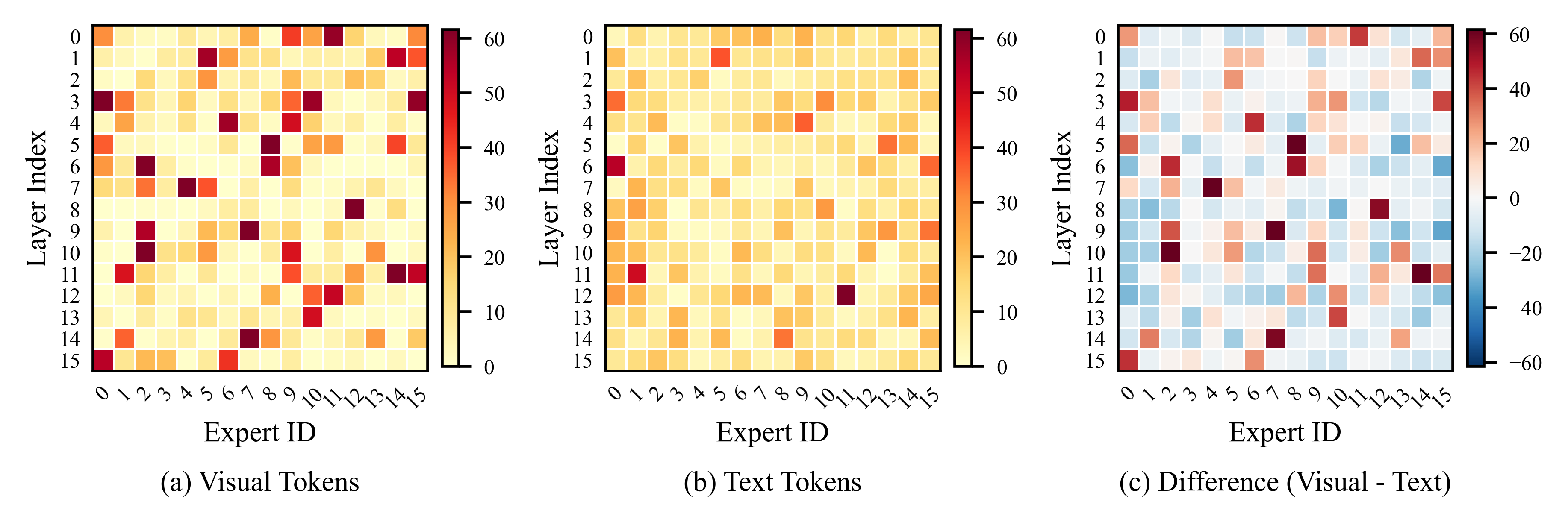}
    \Description[Three heatmaps of modality-specific expert activation.]{Expert activation pattern of MolmoE-1B is shown per-modality. Leftmost and center pane each show visual- and text-centric expert's activation frequency, which are distinct due to their modality-specific characteristics. The rightmost figure shows the subtraction of the two.}
    \vspace{-20pt}
    \caption{Modality-specific expert activation in MolmoE-1B on the COCO caption dataset. (a-b) Each cell shows the percentage of visual/text tokens routed to a given expert (summing to 800 \% per layer due to top-8 routing); darker colors indicate more frequent activations. (c) Difference in percentage points; red and blue mark visual- and text-centric experts, respectively. All three panels share one color scale (0-61.5 \%, $\pm$61.5 \%p in (c)); the first 16 of 64 experts in native index order are shown.}
    \label{fig:fig2}
\end{figure*}

\subsection{Architecture of MoE-based VLMs}
\label{subsec:bg_architecture}

Modern VLMs comprise three principal components: a vision encoder, a modality projector, and a Large Language Model (LLM) backbone. Visual tokens, generated with the vision encoder and projected to the word embedding space via the modality projector, are concatenated with text tokens and forwarded to the LLM backbone for joint processing. As the LLM backbone plays a central role in overall VLM capability, recent efforts have concentrated on scaling it, with increasing adoption of MoE architectures for their compute-efficient scalability. 

As illustrated in Fig. \ref{fig:fig1}, within each MoE layer, the input token $x$ is independently routed to a subset of experts using top-k routing.
The gating function $G(x)$, typically implemented using a linear layer followed by a softmax function, assigns gating weights to each routed expert based on $x$.
The $K$ most relevant experts (i.e., experts with the $K$ highest gating weights) are activated.
In addition, some MoE models employ $M$ shared experts that are always activated.
The output $y$ of the MoE layer combines the weighted sum of outputs from the routed experts with the (unweighted) sum of outputs from the shared experts, formulated as:
\begin{equation} 
    y=\sum^{K}_{i=1}G(x)_{e_i} \cdot E_{e_i}(x) + \sum^{M}_{j=1}S_j(x)
    \label{eq:eq1}
\end{equation}
where $e_i$ is the $i$-th selected expert and $E(x)$ and $S(x)$ denote the output of the routed expert and shared expert, respectively. 

\subsection{Serving Vision-Language Models}
\label{subsec:bg_vlm}

There have been various efforts to enhance the efficiency of VLMs deployed in diverse environments, including production environments \cite{ModServe} or the cloud-edge continuum \cite{ShiftVL}. However, these challenges are exacerbated under the tighter resource constraints of edge deployment.
From an architectural perspective, SparseVLM \cite{SparseVLM} and Focus \cite{Focus} dynamically prune redundant visual tokens, reducing computation overhead while preserving semantic fidelity.
Instead of reducing the number of input tokens, several approaches aimed to reduce the memory overhead at runtime, focusing on Key-Value (KV) cache that grows with the number of tokens processed.
To reduce GPU memory usage, V-Rex \cite{V-Rex} offloads the full KV cache to CPU memory or storage and performs selective retrieval to reduce data transfer latency. To further reduce memory and I/O overhead, Ahn \textit{et al.} \cite{Ahn2026} apply mixed-precision quantization to KV cache, assigning higher bit-widths to important tokens based on attention scores. VLCache \cite{VLCache} avoids costly recomputation by reusing KV cache when the same image is encountered in a subsequent request.
Both the pruning approaches \cite{SparseVLM,Focus} and the work by Ahn \textit{et al.} \cite{Ahn2026} leverage the relative importance of text tokens, which function as semantic anchors for context understanding \cite{Zhang2026,ShiftVL,CEMS}, over spatially redundant visual tokens.
On the other hand, several efforts have reduced model size directly through quantization \cite{MASQuant,MABA,MBQ}, taking into account the separation between text tokens and visual tokens in the latent feature space. 
Still, these quantization-based studies target dense VLMs, leading to suboptimal results when naively applied to MoE-based VLMs.

\subsection{Mixture-of-Experts at the Edge}
\label{subsec:bg_moe}

With the increasing adoption of MoE architecture across various domains, numerous attempts have been made to deploy MoE-based models on-device by alleviating their memory demand.
Many studies have proposed expert-offloading techniques that leverage MoE's sparsely-activated design, storing only frequently activated weights (e.g., attention layers or shared experts) on scarce GPU memory and offloading others to CPU memory or SSD storage, loading them to GPU when activated. Among these, some works perform hybrid CPU/GPU inference to further speed up inference \cite{MoE-Lightning, fiddler, KTransformers}, overlapping GPU-CPU computation and I/O events using pipeline scheduling. However, this approach may not be suitable for devices with limited CPU resources. 
In addition to offloading, several studies quantize experts to reduce their transfer latency and memory footprint. EdgeMoE \cite{EdgeMoE} statically assigns bit-widths to experts based on their impact on model accuracy after being quantized. 
Instead, MoE-APEX \cite{MoE-APEX} and Fate \cite{Fate} dynamically allocate bit-widths to experts based on runtime signals obtained from current inputs: gating weights and the popularity of each expert, respectively.

Separately from offloading, a range of techniques have utilized expert quantization for model compression. MoE-CSP \cite{MoE-CSP}, CMoE \cite{CMoE}, and QMoE \cite{QMoE} quantize experts in uniform precision, down to a minimum of 1 bit, while MC-MoE \cite{MC-MoE} and MxMoE \cite{MxMoE} assign bit-widths based on reconstruction error, each at the expert level and the linear block level.
Still, these approaches target text-only MoE-based LLMs, with only a handful of studies aiming at MoE-based VLMs \cite{MoPEQ, VEQ}.
MoPEQ \cite{MoPEQ} analyzes each expert's importance using Hessian trace approximation and allocates bit-width to each expert group based on the mean importance, yet it neglects the modality-induced disparities across experts.
Meanwhile, VEQ \cite{VEQ} incorporates both expert importance and the sensitivity imbalance between modalities into its error minimization objective. However, its static, uniform quantization lacks flexibility under heterogeneous resource constraints common in edge environments.
In this paper, we combine the modality-specific characteristics of MoE-based VLMs with per-expert activation profiles to adaptively allocate bit-width budgets under per-device resource constraints, jointly addressing what prior studies have left unresolved.

\begin{figure}[t]
    \centering
    \includegraphics[width=\linewidth]{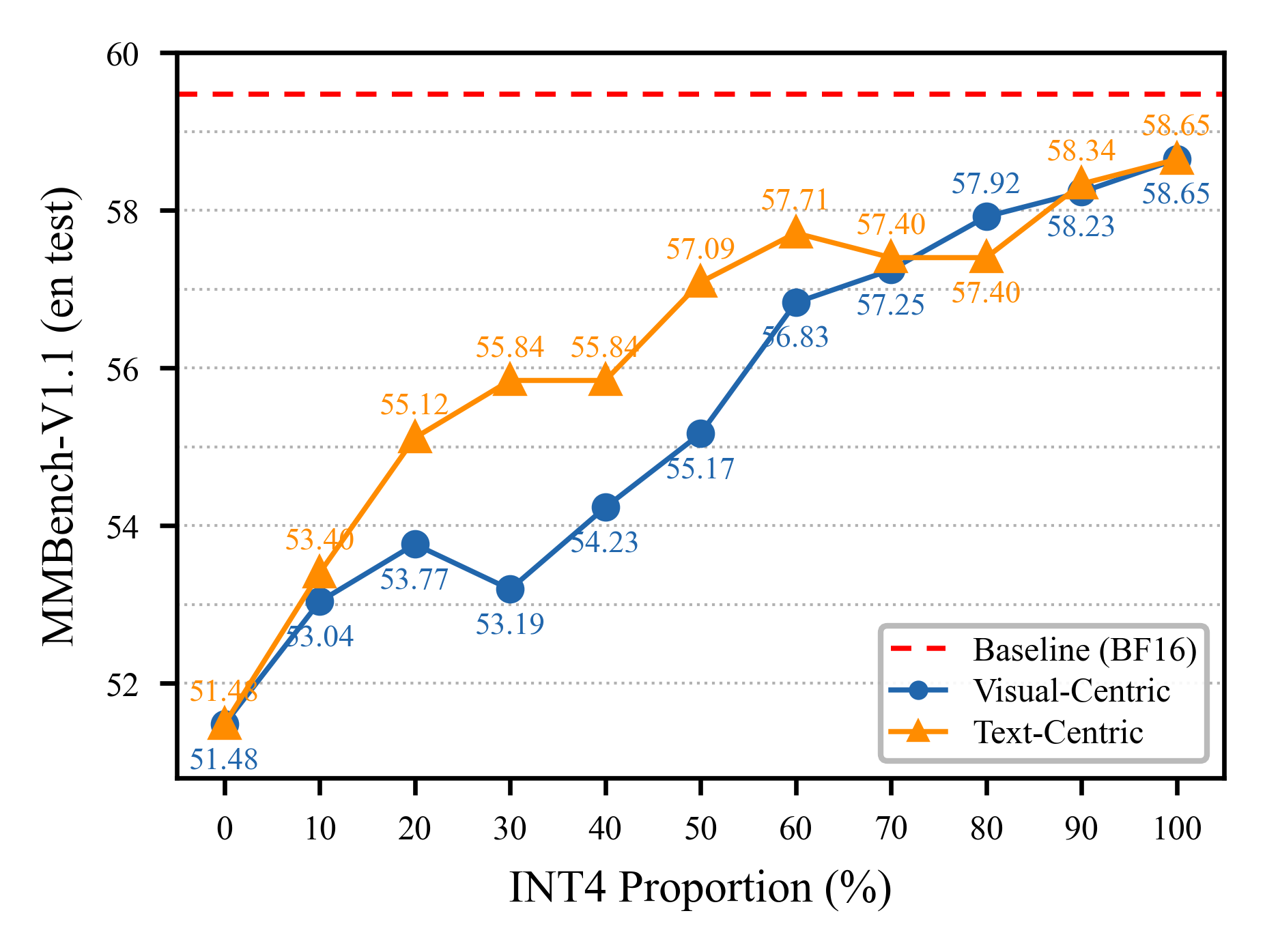}
    \Description[Line plot with orange triangles (Text-Centric), blue circles (Visual-Centric), and a horizontal red dotted line (BF16 baseline)]{Quantizing text-centric experts in INT4 shows better accuracy overall than quantizing visual-centric experts in INT4. Setting INT4 proportion to 30\% shows the largest gap of 2.65.}
    \vspace{-20pt}
    \caption{The trend of benchmark scores when increasing the proportion of experts quantized to INT4 in MolmoE-1B. Visual- and text-centric experts are selected according to their importance when processing visual tokens and text tokens, respectively. Baseline: MolmoE-1B in BF16 (59.48).}
    \label{fig:fig3}
\end{figure}

\section{Insight: Modality-Specific Characteristics of MoE-based VLMs}
\label{sec:insight}
We begin by presenting two key findings from studying modality-induced disparities across experts in terms of activation statistics and quantization sensitivity. Our results demonstrate that modality-specific characteristics of MoE-based VLMs are an effective basis for bit-width assignment. 

\subsection{Imbalanced Activation of Experts Across Modalities}
\label{subsec:imbalanced_activation}

In this section, we analyze the routing behaviors within each MoE layer to measure how the expert activation pattern differs between visual and text tokens.
We use a multimodal calibration task that exercises both the VLM's visual understanding and language modeling capabilities. We choose the improved COCO caption dataset from ShareGPT4V \cite{sharegpt4v} and randomly sample 128 image-caption pairs as the calibration dataset. 
For each visual and text token, we record its top-k index, i.e., the indices of experts selected based on gating weights, and compute the activation frequency of each expert separately for each token modality.

\begin{figure}[t]
    \centering
    \includegraphics[width=0.95\linewidth]{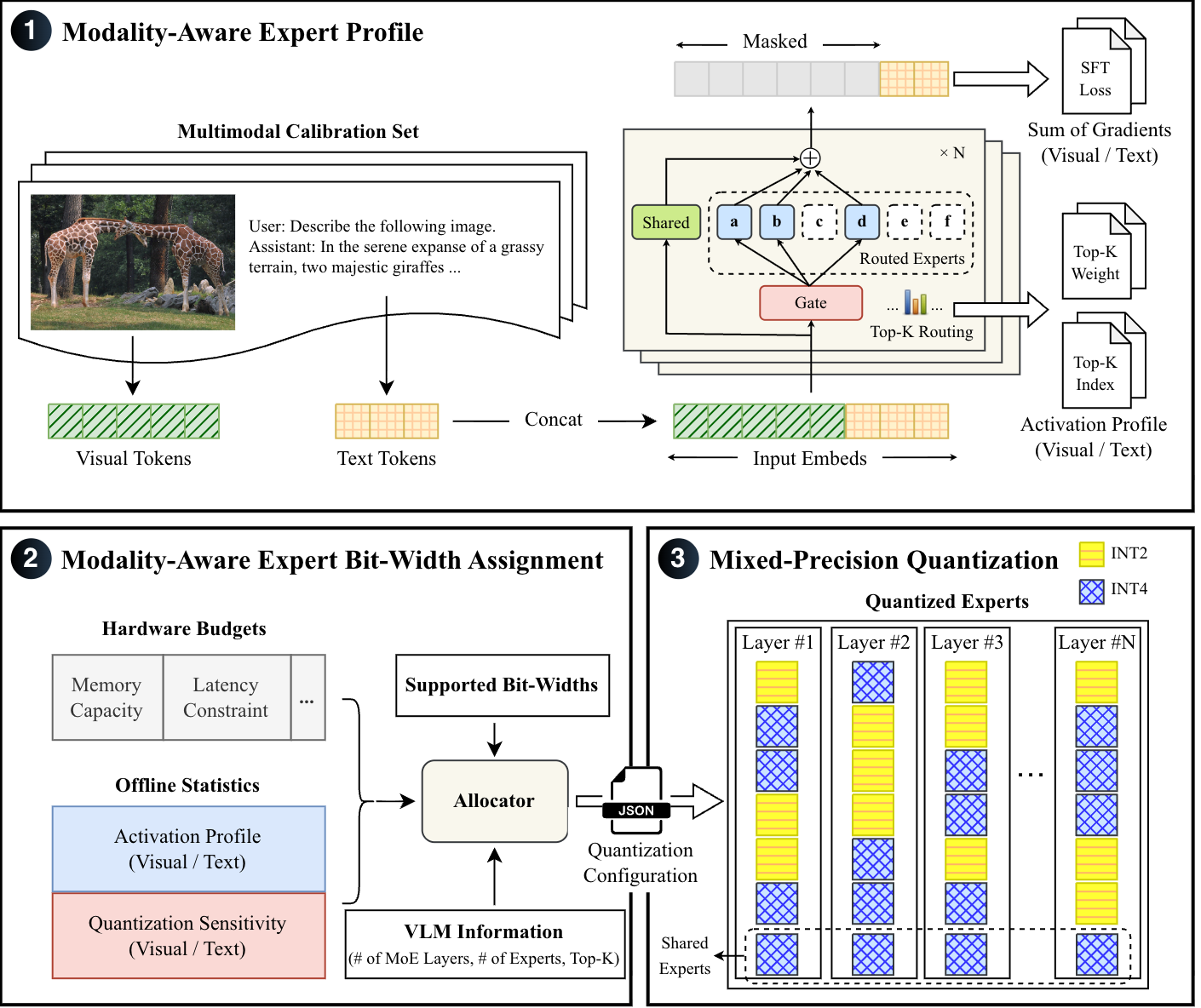}
    \Description[Workflow overview of MoMoVL]{At the first phase, per-expert activation statistics and per-modality quantization sensitivity are measured. At the second phase, the allocator assigns bit-width to each expert based on the collected offline statistics and the provided hardware budgets. At the third phase, mixed-precision quantization is performed on the target model based on the quantization configuration generated at the second phase.}
    \caption{MoMoVL architecture: A workflow overview of an adaptive, modality-aware mixed-precision quantization framework for MoE-based VLMs.}
    \label{fig:fig4}
\end{figure}

Fig. \ref{fig:fig2} shows the modality-specific activation statistics from MolmoE-1B. 
Both modalities demonstrate a skewed activation pattern, while visual tokens exhibit a more concentrated routing than text tokens. When averaged across layers, the 8 most frequently activated experts take 51.49\% of the total activation count when processing visual tokens, while the top 8 take 29.63\% when given text tokens.
This expert specialization diverges by modality. As shown in Fig. \ref{fig:fig2}(c), there is a clear distinction between visual-centric experts and text-centric experts, which are experts frequently activated when processing visual tokens and text tokens, respectively.
Of the 1024 experts across all 16 layers, 61 are activated in more than 50\% when given visual tokens. However, among these 61 visual-centric experts, only 12 also reach even a 25\% activation rate for text tokens, indicating that most visual-centric experts contribute less to processing text.
Still, it is worth noting that there are ``generalist'' experts that are frequently activated on both modalities. At Layer 0, 4 experts (Expert 7, 9, 41, and 43) are activated more than 20\% for both visual and text tokens, an average of 3.5 experts per layer.
Given that visual tokens vastly outnumber text tokens among all input tokens (cf. Table \ref{tab:tab2}), a modality-unaware activation profile would dilute the measured importance of text-centric experts, leading to suboptimal bit-width assignment.
These results indicate that the expert activation profile should be built on a per-modality basis.

\begin{center}
\begin{insightbox}
\texttt{\textbf{Insight 1:}} \textit{Per-modality activation profiles are required to capture the true importance of modality-centric experts.}
\end{insightbox}
\end{center}

\subsection{Modality-Dependent Sensitivity to Expert Quantization}
\label{subsec:quant_sensitivity}

To study the impact of each expert's dominant modality on quantization performance, we select important experts per modality and quantize them to INT4, while quantizing the remaining experts to INT2.
Based on the per-modality activation statistics measured in Sec. \ref{subsec:imbalanced_activation}, we quantize experts in descending order of modality-specific popularity, separately prioritizing either text-centric or visual-centric experts for higher precision.
We use Half-Quadratic Quantization (HQQ) \cite{badri2023hqq}, a fast, calibration-free quantization algorithm that supports various bit-widths. 

\begin{figure}[t]
    \centering
    \includegraphics[width=\linewidth]{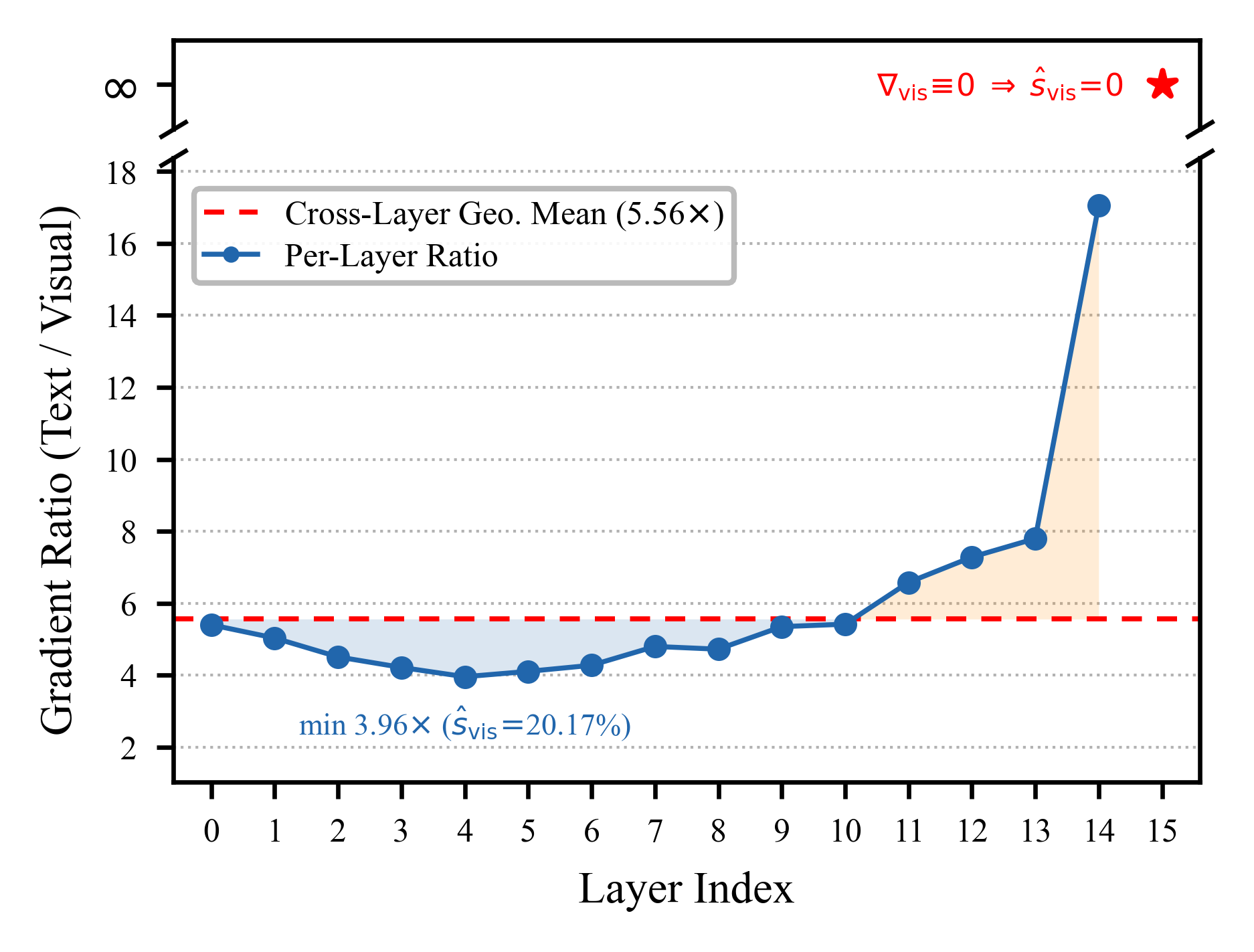}
    \Description[Trend of gradient ratio, text over visual, across layers]{A line plot with blue circles is shown. Ranging from a minimum of 3.96x at Layer 4, then increasing up to Layer 14. The gradient ratio reaches infinity at Layer 15 as visual gradients become 0.}
    \caption{Per-layer ratio of summed absolute gradients of text tokens over visual tokens in MolmoE-1B on the COCO caption dataset. Visual gradients vanish at the last layer due to loss masking and the absence of subsequent attention.}
    \label{fig:fig5}
\end{figure}

Fig. \ref{fig:fig3} shows the result of increasing the proportion of experts quantized to INT4 (i.e., INT4 Proportion) based on per-modality activation frequency. MMBench-V1.1 \cite{liu2024mmbench} is used as the benchmark. While the overall trend shows an increase in accuracy with higher INT4 proportion, it is evident that selectively quantizing text-centric experts to higher bit-width yields higher accuracy than favoring visual-centric experts, with one exception at an INT4 proportion of 80\%.
Note that for both visual-centric and text-centric selection, the accuracy is the same at 0\% and 100\% as all experts are quantized to INT2 and INT4, respectively.
Prioritizing text-centric experts improves accuracy by up to 4.98\% at an INT4 proportion of 30\%.
We attribute this result to the differing quantization sensitivity between text tokens and visual tokens. Text tokens, while fewer in number, carry higher information density and play a central role in a VLM's context understanding and semantic initialization \cite{ShiftVL,MBQ,MABA}.
This, in turn, makes text-centric experts that predominantly process text tokens more susceptible to quantization-induced degradation than visual-centric experts.

\begin{center}
\begin{insightbox}
\texttt{\textbf{Insight 2:}} \textit{Text-centric experts are more susceptible to quanti- zation and deserve higher bit-widths than visual-centric experts.}
\end{insightbox}
\end{center}

\section{MoMoVL Design}
\label{sec:design}
Motivated by our findings on modality-driven imbalance across experts, we propose MoMoVL, a novel modality-aware, adaptive compression framework tailored for on-device MoE-based VLMs, a step toward locally deployed web agents that fit within the memory budgets of edge hardware.

\begin{algorithm}[t]
\caption{Modality-Aware Bit-Width Assignment}
\label{alg:alg1}
\SetAlgoLined
\KwIn{\parbox[t]{0.79\linewidth}{
    Importance metric $M = [M_1, \dots, M_L]$ where \\
    $M_l = [m_{l,1}, \dots, m_{l,N}]$ 
    \hspace{12pt} \texttt{// \textcolor{darkgray}{number of experts $N$}} \\
    Modality scaling factor $S = [\hat{s}_1, \dots, \hat{s}_L]$ where \\
    $\hat{s}_l = (\hat{s}_{l,\text{vis}}, \hat{s}_{l,\text{text}})$ 
    \hspace{17pt} \texttt{// \textcolor{darkgray}{number of MoE layers $L$}} \\
    Options $O = [o_1, \dots, o_K]$, where $o_j = (p_j, r_j)$ pairs bit-width $p_j$ with its budget share $r_j$
}}
\vspace{2pt}
\KwOut{Assignment profile $E = [E_1, \dots, E_L]$ where $E_l = [E_{l,1}, \dots, E_{l,K}]$}
\vspace{2pt}
Sort $O$ in descending order of bit-width \\
\For{$l \gets 1$ \KwTo $L$}{
    $E_l \gets [\emptyset, \dots, \emptyset]$ \tcp*{\textcolor{darkgray}{initialize $K$ empty sets}}
    $C \gets [\,]$ \tcp*{\textcolor{darkgray}{combined importance metric}}
    $C[i] \gets \hat{s}_{l,\text{vis}} \cdot m_{l,i,\text{vis}} + \hat{s}_{l,\text{text}} \cdot m_{l,i,\text{text}}$, \quad $i = 1, \dots, N$ \\
    $T \gets \sum_{i=1}^{N} C[i]$ \\
    $\text{Th} \gets [\,]$ \tcp*{\textcolor{darkgray}{threshold for each bit-width}}
    $\text{Th}[j] \gets T \cdot \sum_{u=1}^{j} r_u$, \quad $j = 1, \dots, K$ \\
    $\pi \gets \arg\text{sort}(C, \text{descending})$ \tcp*{\textcolor{darkgray}{sort experts by $C$}}
    $A \gets 0$; $j \gets 1$ \\
    \For{$i \gets 1$ \KwTo $N$}{
        \While{$A > \text{Th}[j]$ \textbf{and} $j < K$}{
            $j \gets j + 1$ \tcp*{\textcolor{darkgray}{proceed to lower bit-width}}
        }
        $E_{l,j} \gets E_{l,j} \cup \{\pi[i]\}$ \\
        $A \gets A + C[\pi[i]]$
    }
}
\Return $E$
\end{algorithm}

\subsection{Framework Overview}
\label{subsec:framework_overview}
The overall workflow of \momo is depicted in Fig. \ref{fig:fig4}.
\momo first builds an activation profile of each expert separately for visual and text tokens using a small multimodal calibration set (\ding{202}). To measure expert importance, \momo uses one of two types of activation statistics: Top-K Weight or Top-K Index.
Top-K Weight denotes the $k$-highest routing probabilities assigned by the gating function to the $k$ selected experts, given the current input.
As shown in Eq. (\ref{eq:eq1}), the contribution of each routed expert to the output is proportional to its gating weight, i.e., an expert with a higher gating weight has a greater impact on the final output. 
Meanwhile, Top-K Index denotes the indices of the $k$ selected experts. An expert with a higher activation frequency is treated as more important.
We refer to the implementation of \momo based on Top-K Weight as \moW, and the version based on Top-K Index as \moI.
In addition, \momo measures how the impact of quantization noise diverges by modality, using the gradients of the SFT loss to compute a per-layer scaling factor, i.e., the normalized proportion of summed absolute gradients between visual and text tokens. 
\momo reweights each per-modality activation profile by this sensitivity scale, yielding a combined importance metric per expert.
We discuss the measurement of per-modality quantization sensitivity in more detail in Sec. \ref{subsec:measuring_sensitivities}.

Based on the resulting statistics, \momo assigns bit-widths to experts under device-specific budget constraints (\ding{203}). The hardware budget can be formulated generally to support diverse real-world deployment constraints such as memory capacity or inference latency. However, given that memory capacity determines whether a quantized model can be loaded on a target device at all, we focus on memory capacity as the primary constraint in this work.
Details of \momo's expert bit-width assignment are described in Section \ref{subsec:bitwidth_assignment}.
The resulting assignment profile is stored as a JSON file specifying the target bit-width for each expert.
Finally, \momo quantizes the base model according to this quantization configuration, deploying the quantized model on the target device (\ding{204}).

\begin{table}[t]
\caption{Configuration of evaluated MoE-based VLMs.}
\label{tab:tab1}
\centering
\renewcommand{\arraystretch}{1.1}
\newcolumntype{P}[1]{>{\centering\arraybackslash}p{#1}}
\newlength{\tblw}
\setlength{\tblw}{\dimexpr(\columnwidth-6\tabcolsep)\relax}
\begin{tabular}{P{0.4\tblw} P{0.24\tblw} P{0.35\tblw}}
\noalign{\hrule height 1pt}
Metric & MolmoE-1B & DeepSeek-VL2-Tiny \\ 
\hline
\hline
Total Parameters & 7.2B & 3.4B \\
Expert Parameters & 6.4B (89\%) & 2.5B (74\%) \\
\makecell{Experts per Layer\\(Routed / Shared)} & 64 / 0 & 64 / 2 \\
Top-K Routing & 8 & 6 \\
Number of MoE Layers & 16 & 11 \\
\noalign{\hrule height 1pt}
\end{tabular}
\end{table}

\subsection{Measuring Modality-Specific Sensitivities}
\label{subsec:measuring_sensitivities}
In this section, we discuss how \momo measures the per-modality sensitivity.
As shown in Sec. \ref{subsec:quant_sensitivity}, visual tokens and text tokens carry different information density and thus have varying impacts on quantization performance. The question is how to quantify their impact so that we can make a balanced measure of expert importance based on visual- and text-centric activation statistics.
Motivated by MBQ \cite{MBQ} that incorporates gradient-based sensitivity indicators to balance reconstruction loss across modalities, we opt to use the modality-level aggregate of each token gradient derived from the SFT loss as a per-modality sensitivity measure.

As in Sec. \ref{subsec:imbalanced_activation}, we use image-caption pairs from the COCO dataset and apply each model's official chat template to the input image-caption pair to create an instructional format. The SFT loss is computed only from the assistant's response tokens \cite{llava}, while the resulting gradient backpropagates to earlier tokens via causal attention.
For MoE layer $l$, we compute the gradient of the token at position $t$ as its mean absolute value over all hidden dimensions and define the modality-specific sensitivities for a given sample $n$ as the sum of gradients of visual tokens and text tokens, respectively:
\begin{equation}
    g_{l,t} = \frac{1}{d}\sum^{d}_{j=1}\left|\frac{\partial \mathcal{L}}{\partial h_{l,t,j}}\right|
    \label{eq:eq2}
\end{equation}
\begin{equation}
    s^{(n)}_{l,m} = \sum_{t \in {T^{(n)}_{m}}}{g^{(n)}_{l,t}}, 
    \quad
    \bar{s}_{l,m} = \frac{1}{|C|}\sum^{|C|}_{n=1}s^{(n)}_{l,m}, 
    \quad m \in \{\textrm{vis}, \textrm{text}\}
    \label{eq:eq3}
\end{equation}
where $\bar{s}_{l,m}$ denotes the mean of per-sample sums over the calibration set $C$.
These quantities capture how much a small perturbation to each token of modality $m$ would change the final loss, with larger gradients indicating greater sensitivity to quantization noise.
We normalize the two resulting sensitivity measures to sum to 1 as:
\begin{equation}
    \hat{s}_{l,\textrm{vis}} = \frac{\bar{s}_{l,\textrm{vis}}}{\bar{s}_{l,\textrm{vis}}+\bar{s}_{l,\textrm{text}}}, 
    \quad
    \hat{s}_{l,\textrm{text}} = \frac{\bar{s}_{l,\textrm{text}}}{\bar{s}_{l,\textrm{vis}}+\bar{s}_{l,\textrm{text}}}
    \label{eq:eq4}
\end{equation}
These normalized proportions are used as per-layer modality scaling factors, guiding \momo's bit-width assignment.

Fig. \ref{fig:fig5} shows the per-layer ratio of the measured sensitivities, i.e., $\bar{s}_{l,\textrm{text}}/\bar{s}_{l,\textrm{vis}}$, in MolmoE-1B. Overall, text tokens exhibit significantly higher gradients than visual tokens despite being far fewer in number, with the lowest ratio of 3.96$\times$ at Layer 4 and a geometric mean of 5.56$\times$.
Starting from Layer 0, the ratio decreases until Layer 4 and then increases across the remaining intermediate layers, indicating a shift in relative importance between text and visual tokens as layer depth increases. The model relies on the visual modality at shallow layers, but its contribution decreases after the model has extracted the relevant visual information from the input visual tokens. 
However, at the last layer $L$, the visual gradients vanish entirely.
This is because prompt tokens, which include all visual tokens, are excluded from the SFT loss via loss masking and no subsequent attention is performed on the last layer's output.
Consequently, the normalized proportion $\hat{s}_{L,\textrm{vis}}$ becomes 0, indicating that expert importance depends only on the processing of text.
Indeed, at the end of the prefill phase, the first output token is predicted from the last prompt token’s hidden state, which is a text token in correctly formatted prompts.
Moreover, during the decode phase, every input token is a text token generated at the previous forward pass.
This finding leads \momo to assign higher bit-widths exclusively to text-centric experts at the last layer, by design.

\begin{table}[t]
\caption{Statistics of evaluation benchmarks. Token counts are obtained by tokenizing queries with MolmoE-1B.}
\label{tab:tab2}
\centering
\setlength{\tabcolsep}{3pt}
\newcolumntype{P}[1]{>{\centering\arraybackslash}p{#1}}
\begin{tabular}{l@{}r D{.}{.}{4.2} D{.}{.}{3.2} D{.}{.}{2.2}}
\noalign{\hrule height 1pt}
\multirow{2}{*}{Benchmark} 
 & \multirow{2}{*}{\makecell{No. of\\Queries}}
 & \multicolumn{2}{c}{No. of Tokens ($\mu$)} 
 & \multicolumn{1}{c}{\multirow{2}{*}{\makecell{Token Ratio\\(Visual/Text)}}} \\
\cline{3-4}
 & & \multicolumn{1}{c}{Visual} & \multicolumn{1}{c}{Text} & \multicolumn{1}{c}{} \\
\hline
\hline
MMMU (val) & 900 & 831.50 & 129.77 & 6.41\,\mathrm{x} \\
\makecell[l]{MMBench-V1.1\\(en test)} & 7299 & 478.85  & 93.26  & 5.13\,\mathrm{x} \\
SEEDBench & 14232 & 525.35 & 82.73 & 6.35\,\mathrm{x} \\
ScienceQA (test) & 2017 & 487.66  & 52.10  & 9.36\,\mathrm{x} \\
RealWorldQA & 765  & 2050.72 & 98.27  & 20.87\,\mathrm{x} \\
MME-RealWorld (lite) & 1919 & 2623.28 & 176.54 & 14.86\,\mathrm{x} \\
AI2D (test) & 3088 & 792.78  & 80.32  & 9.87\,\mathrm{x} \\
OCRBench & 1000 & 891.90 & 63.30 & 14.09\,\mathrm{x} \\
\noalign{\hrule height 1pt}
\end{tabular}
\end{table}

\begin{table*}[t]
\caption{Main comparison results of MolmoE-1B and DeepSeek-VL2-Tiny across eight multimodal benchmarks. The best results among mixed-precision quantization methods are highlighted in bold. Abbreviations: SEED: SEEDBench, SciQA: ScienceQA, RWQA: RealWorldQA, MME-RW: MME-RealWorld, OCR: OCRBench.}
\label{tab:tab3}
\centering
\renewcommand{\arraystretch}{1.1}
\setlength{\tabcolsep}{3pt}
\newcolumntype{P}[1]{>{\centering\arraybackslash}p{#1}}
\begin{tabular}{c@{\hspace{8pt}}c@{\hspace{8pt}}c@{\hspace{8pt}}c cccccccc}
\noalign{\hrule height 1pt}
Model
 & \makecell{Quantization\\Method}
 & Precision
 & \makecell{Model\\Size (GB)}
 & MMMU
 & MMBench
 & SEED
 & SciQA
 & RWQA
 & MME-RW
 & AI2D
 & OCR \\
\hline
\hline
\multirow{9}{*}{MolmoE-1B} & - & BF16 & 13.45 & 32.22 & 59.48 & 68.57 & 34.80 & 58.69 & 33.87 & 73.41 & 55.0 \\
\cline{2-12}
 & \multirow{2}{*}{\makecell{Uniform\\(Full-Model)}} & INT8 & 7.20 & 32.00 & 59.74 & 68.67 & 34.61 & 58.95 & 33.35 & 73.77 & 55.2 \\
 & & INT4 & 4.08 & 31.11 & 58.18 & 68.06 & 34.41 & 56.73 & 32.57 & 72.96 & 54.6 \\
\cline{2-12}
 & \multirow{2}{*}{\makecell{Uniform\\(Expert-Only)}} & INT4 & 4.45 & 31.89 & 58.65 & 68.19 & 34.31 & 58.30 & 32.31 & 72.99 & 54.9 \\
 & & INT2 & 2.95 & 30.33 & 51.48 & 65.45 & 30.09 & 56.34 & 27.88 & 67.91 & 49.9 \\
\cline{2-12}
 & MoE-APEX & \multirow{4}{*}{\makecell{Mixed\\(INT2, INT4)}} & \multirow{2}{*}{-} & 28.67 & 54.29 & 66.65 & 31.33 & 56.08 & 28.92 & 71.63 & \textbf{53.6} \\
 & Fate & & & 29.56 & 53.56 & 66.27 & \textbf{31.53} & 54.77 & 28.35 & 70.76 & 52.2 \\
\cline{4-12}
 & \shade MoMoVL-Weight & & 3.45 & \textbf{30.67} & \textbf{56.31} & \textbf{67.37} & 30.89 & \textbf{58.56} & \textbf{30.12} & \textbf{71.76} & 52.2 \\
 & \shade MoMoVL-Index & & 3.36 & 29.44 & 56.16 & 67.31 & 31.23 & 58.04 & 29.81 & 71.44 & 52.6 \\
\hline
\multirow{9}{*}{\makecell{DeepSeek\\-VL2-Tiny}} & - & BF16 & 6.30 & 39.44 & 71.58 & 72.38 & 89.59 & 64.97 & 41.58 & 74.71 & 80.7 \\
\cline{2-12}
 & \multirow{2}{*}{\makecell{Uniform\\(Full-Model)}} & INT8 & 3.88 & 39.56 & 71.95 & 72.44 & 89.69 & 64.97 & 40.85 & 75.10 & 80.7 \\
 & & INT4 & 2.67 & 38.78 & 71.22 & 72.10 & 89.14 & 65.36 & 40.80 & 73.22 & 80.3\\
\cline{2-12}
 & \multirow{2}{*}{\makecell{Uniform\\(Expert-Only)}} & INT4 & 2.81 & 39.44 & 71.01 & 72.39 & 89.44 & 64.97 & 41.74 & 74.45 & 80.4 \\
 & & INT2 & 2.25 & 37.11 & 67.43 & 71.40 & 86.76 & 65.23 & 38.09 & 72.22 & 79.0 \\
\cline{2-12}
 & MoE-APEX & \multirow{4}{*}{\makecell{Mixed\\(INT2, INT4)}} & \multirow{2}{*}{-} & 36.89 & \textbf{70.44} & \textbf{72.08} & \textbf{89.64} & 65.36 & 39.71 & \textbf{73.67} & 79.7 \\
 & Fate & & & 37.44 & 70.34 & 71.87 & 88.70 & 65.88 & 38.72 & 73.19 & \textbf{79.9} \\
\cline{4-12}
 & \shade MoMoVL-Weight & & 2.39 & 37.33 & 67.53 & 71.77 & 88.20 & 65.88 & \textbf{40.28} & 72.90 & 78.3 \\
 & \shade MoMoVL-Index & & 2.36 & \textbf{39.11} & 68.31 & 71.63 & 87.61 & \textbf{67.32} & 39.71 & 72.70 & 78.9 \\
\noalign{\hrule height 1pt}
\end{tabular}
\end{table*}

\subsection{Modality-Aware Bit-Width Assignment}
\label{subsec:bitwidth_assignment}

\momo jointly exploits the per-expert activation profile and per-modality quantization sensitivity to derive a quantization configuration tailored for the target device.
Algorithm \ref{alg:alg1} presents the detailed process of \momo's expert bit-width assignment.
Given importance metrics measured separately from visual- and text-centric activation statistics, \momo reweights those metrics by the per-modality scaling factor $\hat{s}_l$, yielding a combined importance metric that captures the true importance of experts. We note that the type of metric can differ by implementation, e.g., using gating weights or activation frequency.

Given a set of bit-widths, each with its own budget share of the total resource budget, \momo ranks experts by this importance score in descending order and processes them in sequence.
\momo assigns each expert the highest available bit-width, proceeding to the next lower bit-width once the accumulated importance exceeds that bit-width's threshold, i.e., its allocated fraction of the total importance score.
As discussed in Sec. \ref{subsec:framework_overview}, the hardware budget and each bit-width's share can be generally instantiated by taking into account diverse deployment challenges. \momo can also assign bit-widths at finer granularity as long as those bit-widths are supported by the quantization algorithm used. In this work, we use INT2 and INT4, which are widely used in expert quantization \cite{Fate,EdgeMoE,MoE-APEX}, and set the fraction of the combined importance metric as the device-specific budget share of each bit-width.

\section{Evaluation}
\label{sec:evaluation}

In this section, we evaluate how \momo enhances on-device vision-language understanding.
Our evaluation is guided by the following three key research questions: 
\begin{itemize}
    \item \textbf{RQ1:} Does \momo improve the accuracy-compression trade-off over state-of-the-art mixed-precision baselines?
    \item \textbf{RQ2:} How do different types of activation statistics affect the accuracy-compression trade-off? 
    \item \textbf{RQ3:} How does each component of \momo contribute to overall performance?
\end{itemize}

\subsection{Experimental Setup}
\label{subsec:eval_setup}

\noindent\textbf{Models.}
We evaluate \momo using two popular MoE-based VLMs: MolmoE-1B \cite{MolmoE} and DeepSeek-VL2-Tiny \cite{DeepSeek-VL2}. As summarized in Table \ref{tab:tab1}, both models employ 64 experts per layer, while DeepSeek-VL2-Tiny additionally uses 2 shared experts that are always activated.
Due to limited GPU resources available, we focus on models under 10B parameters.
We expect \momo to generalize to larger models, since they share the same structural properties and modality-dependent characteristics that motivate our design.

\begin{figure*}[t]
    \centering
    \includegraphics[width=.96\textwidth]{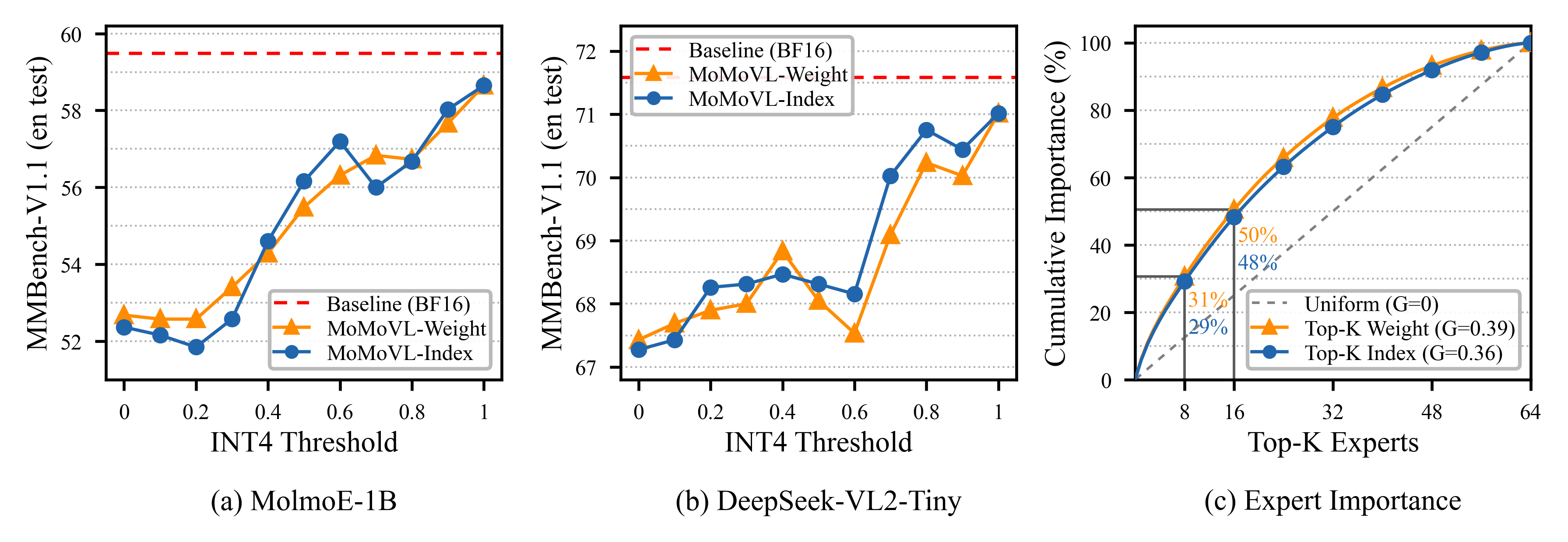}
    \Description[Accuracy-compression trade-off for two models and the distribution of activation statistics]{Both trade-offs for MolmoE and DeepSeek demonstrate that \moI results in higher accuracy when given the same INT4 threshold. This is because activation frequency shows a more balanced distribution than gating weights.}
    \vspace{-10pt}
    \caption{Accuracy-compression trade-off for \momo variants and the skewed distribution of the combined importance metric. (a-b) Benchmark score as the INT4 threshold increases on (a) MolmoE-1B and (b) DeepSeek-VL2-Tiny. (c) Reverse Lorenz curve for MolmoE-1B with experts sorted in descending order of combined importance. Metrics computed from Top-K Weight and Top-K Index show Gini coefficients of 0.39 and 0.36, respectively.}
    \label{fig:fig6}
\end{figure*}

\noindent\textbf{Benchmarks.}
We evaluate \momo on eight widely used public benchmarks: 
MMMU \cite{yue2023mmmu}, MMBench-V1.1 \cite{liu2024mmbench}, SEEDBench \cite{seedbench}, ScienceQA \cite{ScienceQA}, RealWorldQA \cite{RealWorldQA}, MME-RealWorld \cite{MME_RealWorld}, AI2D \cite{ai2d}, and OCRBench \cite{OCRBench}.
These benchmarks encompass diverse tasks spanning multi-discipline reasoning, question answering in real-world and scientific scenarios, and document understanding, which are capabilities central to how web agents interpret visual content.
For each benchmark, we use the standard evaluation protocol and prompts from VLMEvalKit \cite{VLMEvalKit}, an open-source toolkit for evaluating large multimodal models. The details of these evaluation benchmarks are summarized in Table \ref{tab:tab2}.

\noindent\textbf{Baselines.}
We compare \momo with uniform-precision quantization methods at various bit-widths and two state-of-the-art (SOTA) mixed-precision quantization methods to evaluate its efficacy. 
We separately evaluate \moW and \moI to study how the type of activation statistics affects overall performance.
We note that only the LLM backbone is quantized, with all other VLM components kept at their original precision.
For uniform-precision quantization, we quantize either the entire LLM (Full-Model) or only its experts (Expert-Only).

For mixed-precision baselines, we adapt the bit-width assignment logic from MoE-APEX \cite{MoE-APEX} and Fate \cite{Fate}, two SOTA methods that perform mixed-precision quantization on experts based on their importance to accelerate expert loading.
MoE-APEX allocates bit-widths based on gating weights, while Fate utilizes expert popularity estimated by aggregating token-to-expert assignments.
Originally, both methods prefetch experts required at layer $l$+1 by feeding layer $l$'s input into layer $l$+1's gating function, predicting expert importance to assign bit-widths in advance.
To evaluate the performance upper bound of these modality-unaware approaches, we set them in an idealized condition where each layer's expert importance is assessed from that layer's true gating output at runtime.
This isolates the effect of modality-unawareness in their bit-width assignment by removing accuracy loss due to prediction error.
We note that their model sizes are input-dependent due to dynamic precision allocation and are therefore not reported in Table \ref{tab:tab3}.

\noindent\textbf{Quantization Settings.}
For \momo and its baselines, we use HQQ \cite{badri2023hqq} as the quantization algorithm, as discussed in Sec. \ref{subsec:quant_sensitivity}.
\momo is orthogonal to the underlying quantization algorithm and could be paired with other post-training quantization methods.
For \momo and its mixed-precision baselines, experts assigned a higher bit-width are quantized to INT4 while the remaining experts are quantized to INT2. 
Shared experts are fixed at INT4 across all methods.
Fate, following its original implementation, assigns INT4 to all experts during the decode phase.
For a fair comparison, we adopt the original parameters of the baselines as the budget share for 4-bit precision, i.e., INT4 threshold, using 0.6 (MoE-APEX) for \moW and 0.5 (Fate) for \moI.

\subsection{Main Results}
\label{subsec:main_results}
Table \ref{tab:tab3} presents the detailed comparison results on MolmoE-1B and DeepSeek-VL2-Tiny.
For MolmoE-1B, \moW outperforms its mixed-precision baselines on most tasks except for ScienceQA and OCRBench. 
\moI, whose activation statistics (Top-K Index) match Fate's frequency-based mechanism, yields better performance except on MMMU and ScienceQA.
Even though MoE-APEX and Fate dynamically allocate bit-widths to experts based on runtime input at each layer, their modality-unawareness leads to suboptimal results, highlighting the benefit of modality-aware expert importance. 
For DeepSeek-VL2-Tiny, \moI performs competitively with the baselines, achieving the highest MMMU and RealWorldQA scores, though both \momo variants trail the baselines on several benchmarks.
We reason that this result stems from DeepSeek-VL2-Tiny's smaller parameter count, which could leave less capacity to absorb quantization error.

When compared with Uniform (Expert-Only), \momo's overall performance sits between the two extremes, i.e., all experts at INT4 or at INT2, showing the impact of INT4 proportion on model performance.
We discuss this accuracy-compression trade-off and the effect of activation statistics type in Sec. \ref{subsec:acc_comp_tradeoff}.
The performance gap between Uniform (Full-Model) and Uniform (Expert-Only) at INT4 shows the efficiency of expert quantization. 
As experts account for the vast majority of model parameters, extending quantization to non-expert layers results in only an 8.31\% reduction in memory footprint for MolmoE-1B, and 4.98\% for DeepSeek-VL2-Tiny, both accompanied by an overall decrease in performance.
Still, \momo demonstrates its efficacy in model compression. \moW further reduces the memory footprint of Uniform (Full-Model), quantized at INT4, by 15.44\% in MolmoE-1B and 10.49\% in DeepSeek-VL2-Tiny, at the cost of a modest accuracy reduction on most benchmarks, notably outperforming this baseline on RealWorldQA despite its lower average bit-width.

\subsection{Accuracy-Compression Trade-Off}
\label{subsec:acc_comp_tradeoff}
To further study how the type of activation statistics affects \momo's performance, we analyze the change in accuracy when increasing the INT4 threshold, separately for \moW and \moI.
Fig. \ref{fig:fig6} shows the accuracy-compression trade-off and the concentration of expert importance based on Top-K Weight and Top-K Index. We observe that, given an equal budget share for INT4, \moI achieves higher scores than \moW overall, for both MolmoE-1B and DeepSeek-VL2-Tiny.

We attribute this result to the diverging concentration of activation statistics. As discussed in Sec. \ref{subsec:imbalanced_activation}, experts exhibit a skewed activation pattern in terms of activation frequency. This also holds for gating weights. As can be observed in Fig. \ref{fig:fig6}(c), both curves based on Top-K Weight and Top-K Index lie above the uniform diagonal, with the top 25\% of experts covering about half the total importance score. 
Since metrics computed from Top-K Weight are more concentrated than those from Top-K Index, given the same INT4 threshold, i.e., allocated fraction of the total importance score, \moI results in more experts assigned to INT4 than \moW, generally leading to increased accuracy.

It is worth noting that compared with the highly skewed per-modality activation frequency, 
with the top 8 experts accounting for 51.49\% of the total activation count when processing visual tokens, and 29.63\% for text tokens (cf. Sec. \ref{subsec:imbalanced_activation}), the modality-weighted statistics based on Top-K Index show a relatively lower concentration of 29.14\%. This again demonstrates the effectiveness of taking a balanced measure of expert importance by combining per-modality activation statistics.

\begin{table}[t]
\caption{Ablation study of modality-aware expert profile and sensitivity-based scaling on MolmoE-1B, with variants using Top-K Weight (\textit{W}) or Top-K Index (\textit{I}) as importance metric.}
\label{tab:tab4}
\centering
\setlength{\tabcolsep}{3pt}
\newcolumntype{P}[1]{>{\centering\arraybackslash}p{#1}}
\begin{tabular}{l@{}c ccc}
\noalign{\hrule height 1pt}
Variant
 & \makecell{Model\\Size (GB)}
 & MMMU
 & MMBench
 & SEEDBench \\
\hline
\hline
w/o Visual (W) & 3.43 & 30.00 & 55.22 & 67.33 \\
w/o Visual (I) & 3.35 & 30.89 & 55.79 & 67.12 \\
\hline
w/o Modality (W) & 3.27 & 30.33 & 53.71 & 66.43 \\
w/o Modality (I) & 3.21 & 29.56 & 53.35 & 66.29 \\
\hline
w/o Scaling (W) & 3.39 & \textbf{31.11} & 55.43 & 67.22 \\
w/o Scaling (I) & 3.29 & 29.89 & 53.61 & 66.61 \\
\hline
\rowcolor{gray!15} MoMoVL-Weight & 3.45 & 30.67 & \textbf{56.31} & \textbf{67.37} \\
\rowcolor{gray!15} MoMoVL-Index & 3.36 & 29.44 & 56.16 & 67.31 \\
\noalign{\hrule height 1pt}
\end{tabular}
\end{table}

\subsection{Ablation Studies}
\label{subsec:ablation}
We now analyze how each component of \momo contributes to its overall performance with MolmoE-1B.

\subsubsection{Importance of Modality-Aware Expert Profile}
We assess the impact of modality-awareness using text-centric statistics (w/o Visual) and modality-unaware statistics (w/o Modality) as the activation profile.
We omit visual-centric statistics, which underperform text-centric statistics (cf. Sec. \ref{subsec:quant_sensitivity}). 
Table \ref{tab:tab4} shows the evaluation results. 
We observe that both \moW and \moI outperform the ablated versions except for \moI on MMMU. 
When comparing text-centric and modality-unaware versions, solely utilizing text-centric statistics outperforms being modality-unaware in all cases except for using Top-K Weight on MMMU. 
This suggests that while both visual- and text-centric activation profiles are required to capture the true importance of experts, text-centric statistics are more influential among the two.

\subsubsection{Importance of Sensitivity-Based Scaling}
We ablate the impact of sensitivity-based scaling when combining per-modality statistics. 
As shown in Table \ref{tab:tab4}, removing the modality scaling factor, i.e., assigning equal weight to visual- and text-centric statistics, leads to an overall accuracy drop on most tasks except for MMMU.
This result indicates that applying a per-modality scaling factor based on quantization sensitivity leads to a balanced combination between visual- and text-centric activation statistics.

\subsubsection{Impact of Calibration Size}
We analyze how the size of the calibration set affects \momo's performance by varying the number of samples drawn from the same COCO caption dataset.
As shown in Table \ref{tab:tab5}, while both \moW and \moI improve their overall performance as the number of samples increases, their gains saturate after 128. For \moI, we observe that using 128 samples achieves the highest score for all tasks, including a tie on MMMU. This shows that expert activation statistics and per-modality quantization sensitivity converge quickly, requiring relatively little calibration data to measure expert importance.

\begin{table}[t]
\caption{Effect of calibration set size on MoMoVL for MolmoE-1B. Calibration sets are hierarchically nested (i.e., $C_{32} \subset C_{64} \subset C_{128} \subset C_{1000}$) and generated via a fixed random seed.}
\label{tab:tab5}
\centering
\setlength{\tabcolsep}{4pt}
\newcolumntype{P}[1]{>{\centering\arraybackslash}p{#1}}
\begin{tabular}{crccc}
\noalign{\hrule height 1pt}
Method & \makecell{Number of\\Samples}\hspace{3pt}
 & MMMU
 & MMBench
 & SEEDBench \\
\hline
\hline
\multirow{4}{*}{\makecell{MoMoVL\\-Weight}} & 32 & \textbf{31.33} & 56.00 & 67.24 \\
 & 64 & 30.67 & 56.16 & 67.25 \\
 & \shade 128 (default) & \shade 30.67 & \shade 56.31 & \shade 67.37 \\
 & 1000 & 31.00 & \textbf{56.52} & \textbf{67.39} \\
\hline
\multirow{4}{*}{\makecell{MoMoVL\\-Index}} & 32 & 28.89 & 55.74 & 66.96 \\
 & 64 & \textbf{29.44} & 55.90 & 67.19 \\
 & \shade 128 (default) & \shade \textbf{29.44} & \shade \textbf{56.16} & \shade \textbf{67.31} \\
 & 1000 & \textbf{29.44} & 55.79 & 67.17 \\
\noalign{\hrule height 1pt}
\end{tabular}
\end{table}

\section{Conclusion}
\label{sec:conclusion}
In this paper, we proposed \momo, a novel modality-aware, adaptive compression framework tailored for MoE-based VLMs that run under heterogeneous, device-specific resource budgets. 
Built on our empirical findings on modality-induced disparities across experts, \momo jointly exploits the structural properties and modality-dependent characteristics of multimodal MoE models to guide bit-width assignment.
The experimental results demonstrated the benefits of \momo in achieving a better accuracy-compression trade-off over state-of-the-art mixed-precision quantization baselines, and its efficacy in model compression, reducing memory footprint by up to 15.44\% at the cost of a modest accuracy reduction.
By making multimodal inference practical under tight, on-device resource budgets, \momo brings the vision-language reasoning capabilities central to the web agent's perception of visual content within reach of edge deployment.

\begin{acks}
This work was partly supported by the Institute of Information and Communications Technology Planning and Evaluation (IITP) grant funded by the Korea government (MSIT) (No. RS-2023-00234719, SW Star Lab) and the National Research Foundation of Korea (NRF) grant funded by the Korea government (MSIT) (No. RS-2024-00338786). 
\end{acks}

\section*{Ethics and Privacy Statement}
This work presents methodological advances in mixed-precision quantization for MoE-based VLMs, enabling more accurate models to run under tighter on-device resource budgets. 
MoMoVL's offline calibration relies solely on publicly available datasets and requires no access to user data.
Our method does not introduce novel ethical risks beyond those already inherent in multimodal foundation models, such as potential biases from training data and susceptibility to hallucination or misuse.
We note, however, that broadening access to more capable VLMs on low-cost edge hardware could also facilitate undesirable uses, such as unauthorized visual surveillance. 
Still, we believe the benefits of enabling privacy-preserving, on-device multimodal inference for applications such as personal web agents or web accessibility tools outweigh this risk, and we advocate for the responsible use of MoMoVL in benign applications.

\bibliographystyle{ACM-Reference-Format}
\bibliography{References}

\end{document}